\documentclass{article} 
\usepackage{iclr2027_conference,times}

\usepackage{amsmath,amsfonts,bm}

\def\eqref#1{equation~\ref{#1}}

\def\1{\bm{1}}

\DeclareMathAlphabet{\mathsfit}{\encodingdefault}{\sfdefault}{m}{sl}
\SetMathAlphabet{\mathsfit}{bold}{\encodingdefault}{\sfdefault}{bx}{n}

\usepackage{hyperref}
\usepackage{url}
\usepackage{booktabs}
\usepackage{makecell}
\usepackage{graphicx}
\usepackage{float}
\usepackage{subcaption}
\usepackage{cleveref}
\usepackage{booktabs,multirow}
\usepackage{listings}
\hypersetup{hidelinks}

\newtoggle{final}
\iftoggle{final}{
	\usepackage[disable]{todonotes}
}{
    \setlength{\marginparwidth}{1.5cm}
	\usepackage[size=tiny]{todonotes}
}

\iclrfinalcopy

\title{OneDemo: Unified Policy Learning and Data Generation from a Single Demonstration
\thanks{Code available at:
\url{https://github.com/marina-qian/OneDemo}}}

\author{
Ying Qian$^{1}$ \quad
Jianshu Hu$^{2}$ \quad
Han Fang$^{2}$ \quad
Shuguang Li$^{3}$ \quad
Yutong Ban$^{2}$ \\[0.8em]
$^{1}$Stanford University\\
$^{2}$Shanghai Jiao Tong University\\
$^{3}$Tsinghua University
}

\begin{document}

\maketitle
\lhead{Preprint}
\begin{abstract}
{
Learning robot policies from demonstrations is often limited by the cost of collecting diverse, high-quality expert data.
While an appealing alternative is to expand a single successful demonstration into a diverse set of successful trajectories, existing augmentation methods typically rely on manually designed transformation rules or complex trajectory planning, requiring substantial task-specific engineering and human effort. 
We introduce OneDemo, a reinforcement learning framework that \emph{learns to edit a single successful demonstration} for new task configurations.
We formulate demonstration editing as a structured reinforcement learning problem, in which an editor learns, from sparse task-success rewards, both when to transition between object-centric trajectory segments and how to geometrically adapt each segment.
The learned editor serves both as a task policy and as a reusable demonstration generator.} Across 15 Meta-World tasks, OneDemo reaches a 92.7\% success rate within 53.7k environment interactions on average. Across 9 DexMimicGen tasks, it generates successful
demonstrations with a 98.4\% average success rate, and BC-RNN policies trained
on the generated data achieve 99.6\% average success at matched
dataset scale. These results show that learning to edit a single successful demonstration provides a structured and effective basis for both policy learning and scalable demonstration generation.
\end{abstract}

\section{Introduction}

Learning manipulation policies from demonstrations has emerged as a powerful
paradigm for complex robotic tasks, including long-horizon and bimanual
manipulation. Modern imitation learning methods~\citep{zhao2023learning, chi2025diffusion, ze20243d, liu2025rdt} can reproduce sophisticated
behaviors when provided with sufficiently large and diverse demonstration
datasets. Collecting such datasets through human teleoperation, however,
remains expensive and time-consuming, while the resulting demonstrations
typically cover only a small fraction of the task configurations a robot may
encounter.

A natural alternative is to generate additional demonstrations by adapting a
small number of successful trajectories to new object poses, scene
configurations, or task initializations.
If these adaptations can be performed
reliably, a single demonstration can provide supervision over a substantially
broader task distribution.
Existing approaches toward this goal face an
important trade-off.
Simple perturbation-based augmentation 
is easy to apply~\citep{wang2024cyberdemo,zhou2023nerf},
but can produce invalid or unsuccessful trajectories when the target task configuration
deviates substantially from the source demonstration.
More structured approaches, such as MimicGen \citep{mandlekar2023mimicgen}, obtain substantially
higher reliability by decomposing demonstrations into object-centric segments
and geometrically transforming them to new scenes.
Their behavior, however,
is largely determined by manually specified task structure and trajectory-generation procedures.

\begin{figure}[t]
    \centering
    \includegraphics[width=1.0\textwidth]{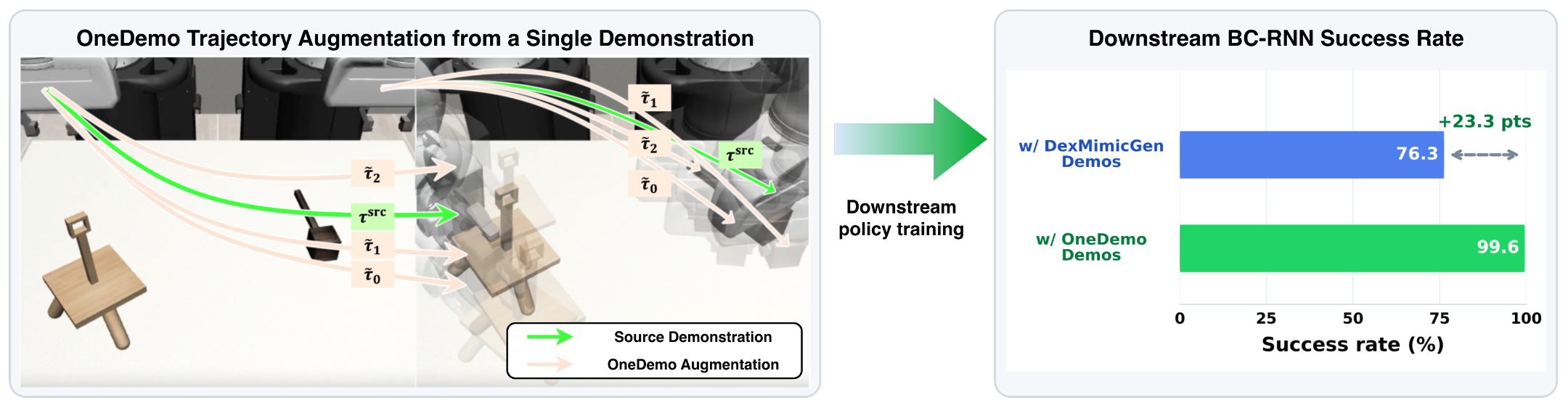}
    \caption{
    \textbf{OneDemo learns to edit a single 
    demonstration to generate {varying} successful trajectories 
    {under} new spatial configurations.}
    \textbf{Left:} OneDemo adapts the source trajectory to generate successful augmentations.
    \textbf{Right:} Recurrent behavior cloning (BC-RNN) trained on 
    {demonstrations augmented by OneDemo} improves success from $76.3\%$ to $99.6\%$ compared with DexMimicGen data.
    }
    \label{fig:teaser}
\end{figure}

In this work, we ask whether the edits required to adapt a successful
demonstration to a new task configuration can instead be \emph{learned}.
Recent work has begun to explore related directions: DemoGrasp ~\citep{yuan2026demograsp} learns a
single global edit to a demonstrated dexterous grasp using reinforcement
learning, while DexFlyWheel ~\citep{zhu2025dexflywheel} combines demonstration-based augmentation with residual reinforcement learning in an iterative data-generation pipeline.
Building on this emerging direction, we investigate a more structured form of demonstration editing that extends to long-horizon, bimanual, and dexterous manipulation.

Starting from a single source trajectory, we formulate demonstration editing as
a structured reinforcement learning problem. Rather than exploring directly over
low-level robot actions, the policy operates over edits to an existing successful
motion: it learns when to transition between object-centric trajectory segments, and how to apply compact geometric residual corrections to their motion.
Specifically, we represent the source demonstration as an ordered sequence of object-centric stages, each associated with an anchor object or frame.
At the beginning of each stage, the policy predicts both a duration, which
determines the number of timesteps assigned to that stage, and a
rigid-body residual transformation that adapts the corresponding motion to the
current scene.
We optimize these decisions using PPO~\citep{schulman2017proximal} with only sparse task-success rewards, with the anchor sequence providing only a coarse initialization.

A key consequence of this formulation is that the learned trajectory editor serves both as a task policy and as a demonstration generator. As a task policy, it adapts and executes the source demonstration under new task configurations, while restricting exploration to structured modifications of an already successful trajectory rather than low-level robot actions. This structured exploration can substantially simplify the reinforcement learning problem. The same editor can then be rolled out across task configurations to generate successful trajectories for downstream learning.

Our experiments evaluate both roles of the proposed system.
When deployed directly as a policy, OneDemo learns
high-performing task behavior from a single source demonstration within a
small interaction budget.
When used as a demonstration generator, it produces high-quality trajectories that substantially improve downstream imitation learning at matched dataset scale.

Our main contributions are:
\begin{enumerate}
    \item We introduce \textsc{OneDemo}, a structured reinforcement learning
    framework that uses a single successful demonstration to define a compact
    search space for adapting behavior to new spatial task configurations.

    \item We formulate demonstration adaptation as learning two complementary
    edits from sparse task-success rewards: temporal segmentation of the source
    trajectory and object-centric residual transformations that geometrically
    adapt each trajectory segment to the current task configuration.

    \item We show that the learned trajectory editor serves both as a sample-efficient task policy and as a high-quality demonstration generator. When used directly as a policy, it achieves strong performance with only a small budget of online environment interactions. When used for data generation, it produces successful demonstrations across long-horizon, bimanual, and dexterous manipulation tasks that substantially improve downstream policy learning.
\end{enumerate}

\section{Related Work}
\paragraph{Imitation Learning.}
Imitation learning aims to learn policies from demonstrations of desired behavior~\citep{osa2018algorithmic}. Common approaches include behavior cloning with recurrent policies~\citep{mandlekar2021matters}, action-chunking methods such as ACT~\citep{zhao2023learning}, and diffusion-based policies~\citep{chi2025diffusion,ze20243d}. While these methods differ in architecture and prediction strategy, their performance depends strongly on the coverage and diversity of the demonstration dataset~\citep{mandlekar2021matters,jiang2025dexmimicgen}. Interactive methods such as DAgger~\citep{ross2011reduction} and human-in-the-loop correction~\citep{mandlekar2020human} improve coverage by collecting expert supervision on states encountered by the learned policy, but typically assume continued access to an expert. Our system can be seen as a learned expert policy from a single source demonstration, which can both generate diverse offline training trajectories and potentially provide supervision within interactive imitation-learning schemes such as DAgger.

\paragraph{Reinforcement Learning with Demonstrations.}
Reinforcement learning can learn task policies without expert demonstrations, but exploration becomes difficult for long-horizon manipulation tasks with sparse rewards. To address this problem, prior work improves sample efficiency by using demonstrations to initialize policies~\citep{rajeswaran2017learning}, populate replay buffers~\citep{vecerik2017leveraging}, provide auxiliary imitation losses~\citep{rajeswaran2017learning, nair2018overcoming}, or to initialize episodes from favorable states~\citep{nair2018overcoming}. Residual RL~\citep{johannink2019residual}, similarly, learns corrections on top of an existing policy to restrict policy search space. Our method also uses demonstrated behavior as a strong prior, but directly learns how the demonstrated trajectory itself can be modified to generalize to new task configurations.

\paragraph{Synthetic Demonstration Generation.}
Another line of work reduces teleoperation cost by generating additional trajectories from a small set of source demonstrations. Simple perturbation-based methods~\citep{wang2024cyberdemo,zhou2023nerf} introduce variations around existing trajectories but may produce mostly unsuccessful rollouts. More structured methods such as MimicGen~\citep{mandlekar2023mimicgen} decompose demonstrations into object-centric segments and transform them geometrically to match new scene configurations, enabling much more reliable synthesis. Such approaches~\citep{jiang2025dexmimicgen, Garrett2025skillmimic}, however, rely on task-specific structure and transformation rules. Our work instead learns to do trajectory adaptation through reinforcement learning, allowing successful transformations to emerge from interactions rather than being explicitly prescribed based on prior knowledge. 

\section{Method}
\label{sec:method}
Given a single successful source demonstration, our goal is to learn to adapt
it to new spatial configurations of the same task using only sparse
task-success rewards. In this work, we define a new task configuration as one in which the
object geometry and task requirements remain fixed, while the poses of
task-relevant objects vary.
To make this adaptation practical for long-horizon tasks and tasks requiring
high spatial precision, we avoid reinforcement learning directly over
low-level robot actions. Such exploration can require hundreds of coordinated decisions before receiving a sparse success
signal. Instead, we restrict exploration to structured edits of an existing
successful trajectory.

\begin{figure*}[t]
    \centering
    \includegraphics[width=\textwidth]{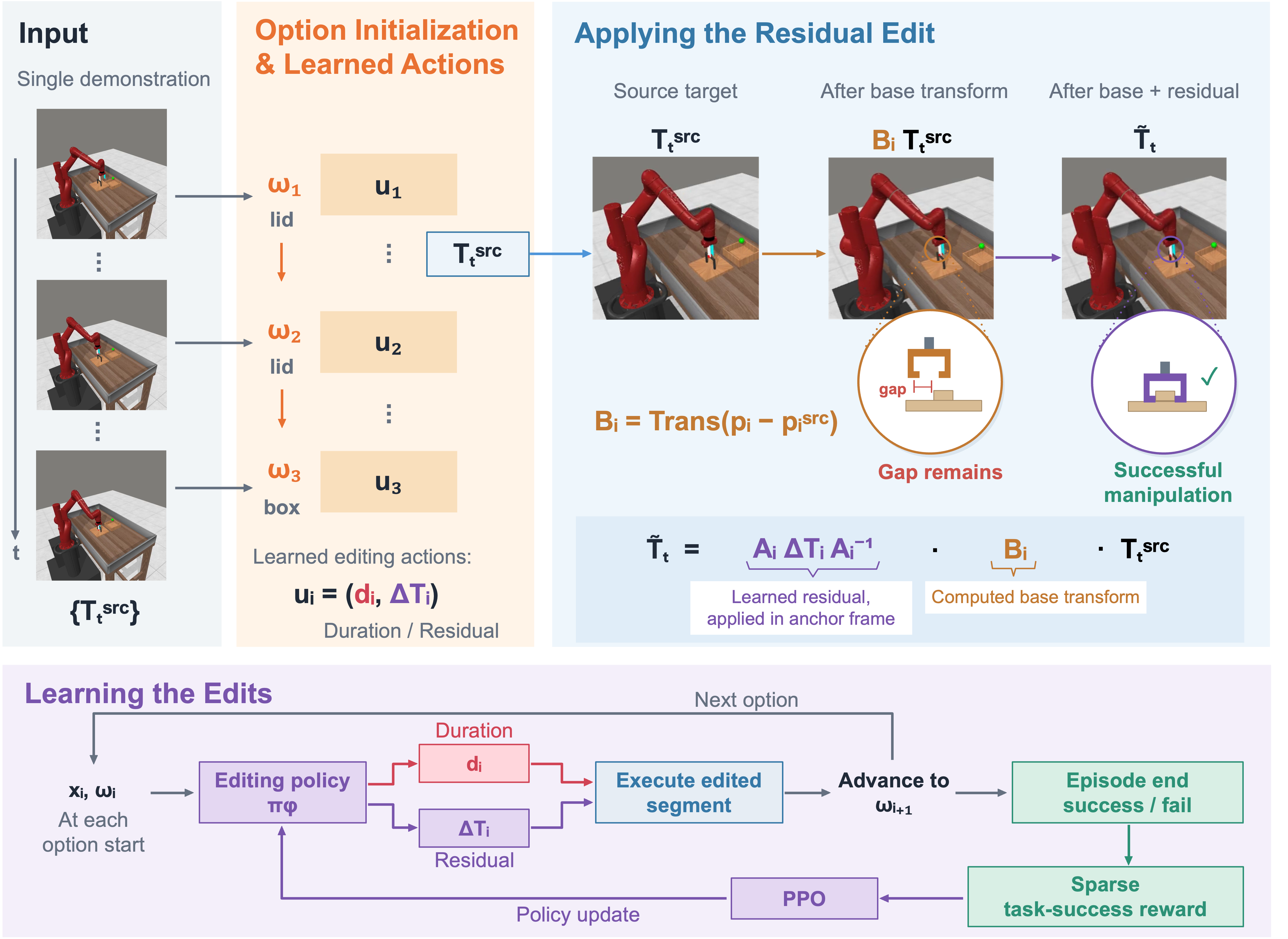}
    \caption{
    \textbf{Overview of OneDemo}. Semantic options define anchor objects and base transform, while the learned policy predicts an option duration and residual transformation at each option boundary. The residual is applied on top of the base transform and optimized from sparse task-success rewards.
    }
    \label{fig:method_overview}
\end{figure*}

Our method represents the source demonstration as an ordered sequence of object-centric trajectory segments.
For each segment, the policy decides \emph{when the segment should end} and \emph{how its motion should be geometrically adapted} to the current task configuration.
The resulting trajectory is then executed by the {low-level controller of the robot.}
We first formulate this representation as a structured specialization of the options framework {(\Cref{sec:structured_options,sec:trajectory_editing})}, and then describe its realization and initialization {(\Cref{sec:option_policy})}.
Figure~\ref{fig:method_overview} provides an overview of the full pipeline.

\subsection{Option-Based Formulation}
\label{sec:structured_options}
We first present the formulation for a single manipulator, where the standard call-and-return interpretation of options applies directly. We describe the bimanual extension used in our experiments later in \Cref{par:bimanual_ext}.

Our formulation builds on the options framework~\citep{sutton1999between}, which augments a Markov decision process with temporally extended actions. We provide a brief overview in Appendix~\ref{app:options}. A Markovian option $\omega$ is conventionally defined by
\[
\omega = (I_\omega,\pi_\omega,\beta_\omega),
\]
where $I_\omega$ is its initiation set, $\pi_\omega$ is the intra-option policy, and $\beta_\omega$ determines when the option terminates. Option-Critic~\citep{bacon2017option} showed that the intra-option policy and termination mechanism can be optimized jointly from task return, without requiring option-specific
rewards.

We specialize this abstraction to demonstration adaptation. We assume an ordered sequence of semantic options
\[
\omega_1 \rightarrow \omega_2 \rightarrow \cdots \rightarrow \omega_K.
\]
When $\omega_i$ terminates, execution advances deterministically to
$\omega_{i+1}$, so no policy over options needs to be learned. The remaining option components correspond to two trajectory-editing decisions:
\[
\beta_\omega \;\mapsto\; d_i
\quad\text{(transition timing)},\qquad
\pi_\omega \;\mapsto\; \Delta T_i
\quad\text{(trajectory adaptation)}.
\]
At the start of option $\omega_i$, the intra-option policy predicts both a duration $d_i$ and a rigid-body transformation $\Delta T_i$ applied to the corresponding source-demonstration segment. Thus, each option-level action is
\[
u_i=(d_i,\Delta T_i).
\]

Since $u_i$ persists for a variable number of primitive control steps, the
resulting option-level process is naturally semi-Markov~\citep{sutton1999between}.
This preserves the temporal abstraction while restricting exploration to structured trajectory edits rather than low-level robot actions.

\subsection{Object-Centric Trajectory Editing}
\label{sec:trajectory_editing}

Given a task, let a successful source demonstration $\tau^{\mathrm{src}}$ contain a
sequence of end-effector target poses:
\[
\tau^{\mathrm{src}} = \{T_t^{\mathrm{src}}\}_{t=0}^{H-1},
\qquad
T_t^{\mathrm{src}} \in SE(3),
\]
where $T_t^{\mathrm{src}}$ denotes the end-effector target pose at time step $t$. 
As introduced above, we represent the task as an ordered sequence of semantic options {$\Omega$:} \[ \Omega = (\omega_1,\ldots,\omega_K), \] where each option $\omega_i$ is associated with an anchor object or task frame corresponding to a particular stage of the manipulation.

Our key geometric assumption is that, although relevant object poses vary across task configurations, successful motion relative to those objects is approximately reusable. For option $\omega_i$, let $A_i^{\mathrm{src}}$ denote the pose of its associated
anchor object in the source demonstration, and let $p_i^{\mathrm{src}}$ be the translation
component of $A_i^{\mathrm{src}}$. Correspondingly, let $A_i$ denote the live pose of the same anchor during execution, with translation component $p_i$.

For a given task configuration, we initialize trajectory adaption for option $\omega_i$ using translation-only anchor alignment,
\[
B_i
=
\operatorname{Trans}(p_i-p_i^{\mathrm{src}}).
\]
The editing policy then predicts an additional residual transformation
\[
\Delta T_i \in SE(3),
\]
expressed in the live anchor frame and held fixed throughout the option.

Let $\mathcal{T}_i$ denote the source-trajectory indices assigned to
$\omega_i$. For $t\in\mathcal{T}_i$, the edited target is
\[
\widetilde T_t
=
A_i \Delta T_i A_i^{-1}
B_i T_t^{\mathrm{src}}.
\]
Only the end-effector pose trajectory is geometrically edited. Gripper and
dexterous-hand commands are replayed unchanged from the source demonstration.
Here, $B_i$ provides a coarse adaptation to the new anchor location, while
$\Delta T_i$ learns the remaining geometric correction required for successful
execution.

This construction is related to the object-centric trajectory transformations
used by MimicGen~\citep{mandlekar2023mimicgen} and DexMimicGen~\citep{jiang2025dexmimicgen}. The key distinction is that OneDemo
treats the temporal segmentation and residual geometric adaptation as learned decisions rather than prescribing them in advance. Additional motivation for
our translation-only initialization is provided in
Appendix~\ref{app:anchor_alignment}.

\paragraph{Bimanual Extension.}
\label{par:bimanual_ext}
We extend the single-arm formulation to the bimanual setting by maintaining a separate ordered semantic phase sequence and
anchor list for each arm. Each arm has its own active phase, duration, and held residual
warp, while both arms execute on the same synchronized control clock. Phase
boundaries may occur asynchronously across arms.

\subsection{Learning Temporal Boundaries and Geometric Adaptations}
\label{sec:option_policy}

Suppose option $\omega_i$ begins at source-trajectory time index $\kappa_i$.
At option initiation, the policy $\pi$, parameterized by $\phi$, samples the
option-level action
\[
u_i = (d_i,\Delta T_i)
\sim
\pi_\phi(\cdot\mid x_i,\omega_i),
\]
where $x_i$ denotes the policy observation at the beginning of the option.
The duration $d_i$ determines the source-trajectory segment assigned
to option $\omega_i$:
\[
\mathcal{T}_i
=
\{\kappa_i,\ldots,\kappa_i+d_i-1\},
\qquad
\kappa_{i+1}=\kappa_i+d_i.
\]
The residual transformation $\Delta T_i$ is held fixed while the transformed
targets indexed by $\mathcal{T}_i$ are executed. Once this segment is completed,
execution advances to the next option $\omega_{i+1}$.

Conditioned on the active option, we factorize the editing policy as
\[
\pi_\phi(d,\Delta T\mid x,\omega)
=
\pi_{\vartheta_\omega}(d\mid x)
\,
\pi_{\theta_\omega}(\Delta T\mid x),
\]
where $\vartheta_\omega$ parameterizes the duration distribution and
$\theta_\omega$ parameterizes the residual transformation distribution for
option $\omega$. The complete policy parameters are
\[
\phi
=
\{\vartheta_\omega,\theta_\omega\}_{\omega\in\Omega}.
\]

We optimize the expected task return
\[
J(\phi)
=
\mathbb{E}_{\pi_\phi}[R]
\]
using PPO, where $R$ is the sparse task-success return. Both temporal
segmentation and geometric adaptation are therefore learned from end-task
performance only.

The resulting search space contains only a small number of structured
option-level decisions per episode,
\[
\{u_i\}_{i=1}^{K},
\]
rather than a long sequence of unconstrained low-level actions. This reduction
is particularly useful for manipulation tasks in which successful trajectories
require coordinated motion over long horizons but remain locally reusable
relative to task-relevant objects.

\paragraph{Automatic Option Initialization.}
\label{par:option_init}
OneDemo requires an ordered sequence of semantic options and associated
task-relevant anchors for initialization. We obtain this sequence automatically
using an LLM given task metadata and downsampled telemetry from the source
demonstration. The LLM predicts only a coarse semantic decomposition and the
anchor associated with each option; precise temporal boundaries and geometric
corrections are learned by OneDemo from task return. The initialization need
not be minimal, since redundant options can be effectively shortened by the
learned duration policy. Full prompting details are provided in
Appendix~\ref{app:option_initialization}.

\section{Experiments \& Results}
We evaluate OneDemo along two complementary roles, and perform additional ablation
studies that analyze its key components:
\begin{enumerate}
    \item \textbf{OneDemo as a policy:} We study whether the structured
    trajectory-editing action space enables sample-efficient reinforcement
    learning when OneDemo itself is deployed as the task policy.

    \item \textbf{OneDemo for trajectory generation:} We evaluate whether the
    trajectories generated by OneDemo provide high-quality supervision for
    downstream policy learning.

    \item \textbf{Ablation studies:} We analyze the respective contributions
    of learned temporal segmentation and geometric residual adaptation to option-level policy learning.
\end{enumerate}

\subsection{Experimental Setup}
\label{sec:implementation_details}
We evaluate OneDemo primarily on two manipulation benchmarks spanning
single-arm, bimanual, and dexterous manipulation, with additional
complementary results on RoboTwin. Each benchmark is used to
evaluate a different aspect of the method. Full implementation details,
network architectures, and training hyperparameters are provided in
Appendix~\ref{app:implementation}.

\paragraph{Meta-World.}
Meta-World~\citep{yu2020meta} contains single-arm manipulation tasks with
a parallel-jaw gripper under randomized task configurations. We evaluate on
15 challenging tasks matching the subset used by MoDem~\citep{hansen2022modem},
a demonstration-guided model-based RL method designed for high sample efficiency
under sparse rewards. We use this benchmark to evaluate OneDemo directly as a task policy, considering both the success rate of the resulting policy and the number of online environment interactions required for training.

\paragraph{DexMimicGen.}
DexMimicGen~\citep{jiang2025dexmimicgen} contains 9 complex bimanual manipulation tasks
spanning three robot embodiments, including gripper-based, dexterous-hand, and
humanoid manipulation. We use this benchmark to evaluate OneDemo as a
demonstration generator. For each task, we generate a new dataset from a single
source demonstration and evaluate its quality through the performance of
downstream BC-RNN policies trained on the generated trajectories.

\paragraph{Additional Benchmark.}
We provide complementary results on RoboTwin~\citep{chen2025robotwin} in the appendix. On selected bimanual tasks, we evaluate OneDemo-generated data in a low-data regime using Diffusion Policy trained from only 50 generated trajectories per task.

\subsection{OneDemo as a Sample-Efficient Policy}
\label{sec:exp_policy}
We first evaluate OneDemo directly as a task policy on the 15 Meta-World tasks used by MoDem~\citep{hansen2022modem}. For each task, we initialize from a single successful scripted demonstration and train using only sparse task-success rewards. We compare against our local reproduction of MoDem, a demonstration-guided model-based RL method, to contextualize the interaction budget required for sparse-reward policy learning. Additional details on the baseline and comparison protocol are provided in Appendix~\ref{par:modem_baseline}.

\paragraph{Sample Efficiency.}
\label{sec:metaworld_efficiency}
Table~\ref{tab:metaworld_results} summarizes the results. 
Under a budget of $200$k environment interactions, OneDemo achieves an average best success rate of $92.7\%$ and reaches its best checkpoint after only $53.7$k interactions on average.

\begin{table}[t]
    \centering
    \caption{\textbf{Meta-World success rate on the 15 tasks used by MoDem.}
    \textbf{Naive} uses the randomly initialized duration policy with greedy
    sampling and applies only the baseline transformation, with no residual warping.
    \textbf{Seg.-only} uses the learned duration policy with no residual warping.
    \textbf{OneDemo} uses both the learned duration policy and learned residual warping.
    Values are mean $\pm$ 95\% confidence-interval half-width over five seeds;
    each evaluation uses 50 episodes.}
    \label{tab:metaworld_results}

    \small
    \setlength{\tabcolsep}{8pt}
    \renewcommand{\arraystretch}{0.96}
    \begin{tabular}{@{}lccc@{}}
        \toprule
        \textbf{Task} &
        \textbf{Naive} &
        \textbf{Seg.-only} &
        \textbf{OneDemo} \\
        \midrule
        Assembly        & $46.4 \pm 42.9$ & $\mathbf{100.0 \pm 0.0}$ & $\mathbf{100.0 \pm 0.0}$ \\
        Basketball      & $68.8 \pm 13.2$ & $76.8 \pm 5.2$           & $\mathbf{84.4 \pm 5.5}$ \\
        Box Close       & $57.6 \pm 8.5$  & $66.8 \pm 13.5$          & $\mathbf{83.6 \pm 14.2}$ \\
        Coffee Push     & $64.8 \pm 41.7$ & $99.2 \pm 1.0$           & $\mathbf{100.0 \pm 0.0}$ \\
        Hand Insert     & $77.2 \pm 22.7$ & $99.2 \pm 1.6$           & $\mathbf{100.0 \pm 0.0}$ \\
        Peg Insert Side & $43.6 \pm 16.4$ & $\mathbf{98.4 \pm 3.1}$  & $\mathbf{98.4 \pm 3.1}$ \\
        Pick Place      & $89.6 \pm 20.4$ & $\mathbf{100.0 \pm 0.0}$ & $\mathbf{100.0 \pm 0.0}$ \\
        Pick Place Wall & $93.2 \pm 13.3$ & $\mathbf{100.0 \pm 0.0}$ & $\mathbf{100.0 \pm 0.0}$ \\
        Push            & $82.0 \pm 34.3$ & $\mathbf{100.0 \pm 0.0}$ & $\mathbf{100.0 \pm 0.0}$ \\
        Push Wall       & $47.6 \pm 31.4$ & $92.8 \pm 4.7$           & $\mathbf{97.6 \pm 2.9}$ \\
        Soccer          & $60.0 \pm 19.8$ & $73.6 \pm 5.0$           & $\mathbf{75.2 \pm 4.6}$ \\
        Stick Pull      & $22.8 \pm 15.3$ & $45.2 \pm 7.1$           & $\mathbf{56.8 \pm 12.7}$ \\
        Stick Push      & $\mathbf{100.0 \pm 0.0}$ & $\mathbf{100.0 \pm 0.0}$ & $\mathbf{100.0 \pm 0.0}$ \\
        Sweep           & $68.4 \pm 26.9$ & $\mathbf{100.0 \pm 0.0}$ & $\mathbf{100.0 \pm 0.0}$ \\
        Sweep Into      & $87.6 \pm 8.5$  & $94.0 \pm 4.5$           & $\mathbf{95.2 \pm 4.4}$ \\
        \midrule
        \textbf{Average} & 67.3 & 89.7 & \textbf{92.7} \\
        \bottomrule
    \end{tabular}
\end{table}

\begin{figure}[t]
    \centering
    \includegraphics[width=1.0\columnwidth]{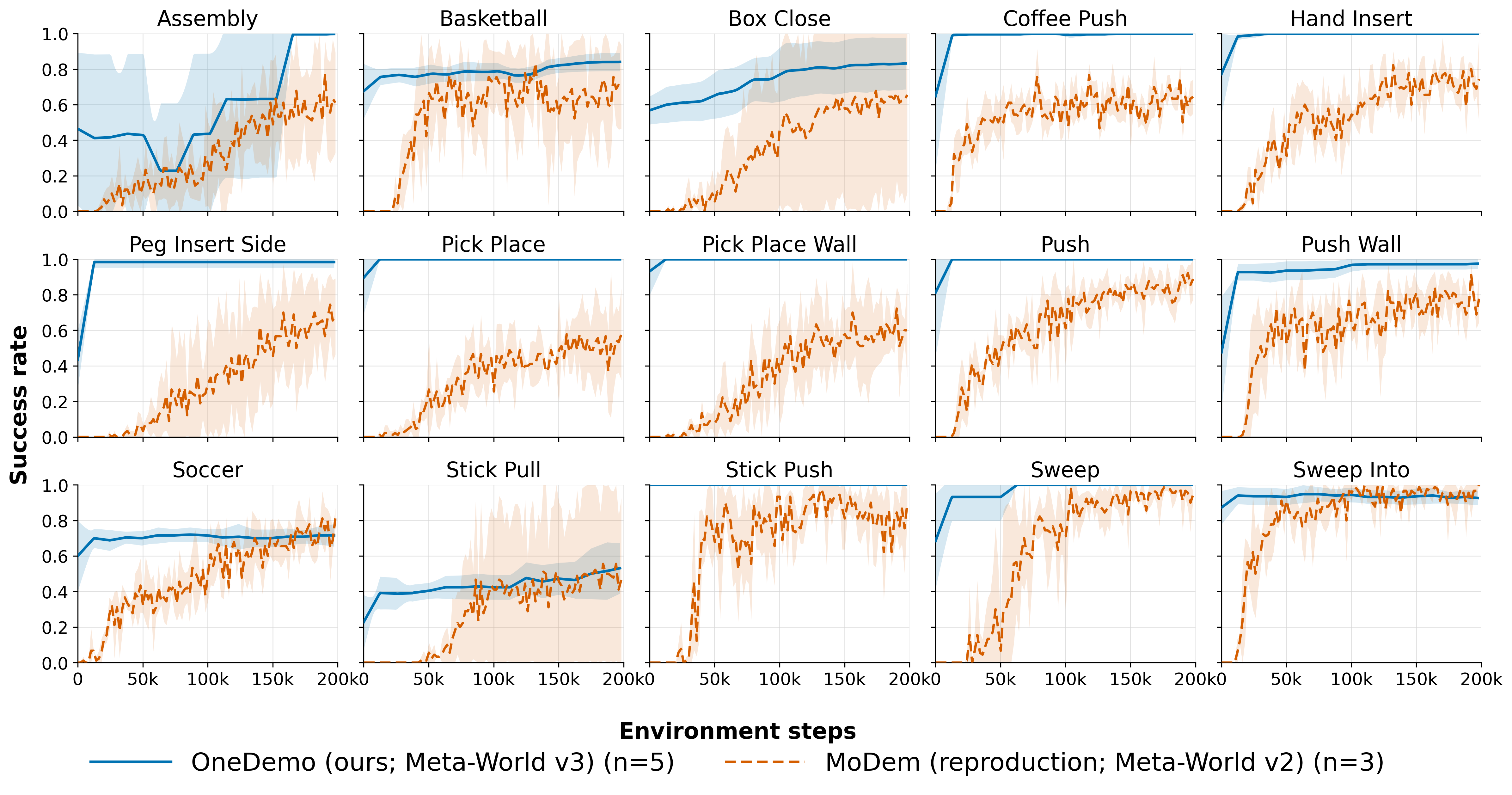}
    \caption{
    \textbf{Meta-World sample efficiency aggregated over 15 tasks.}
    OneDemo directly executes the edited demonstration as the task policy and
    achieves high success with substantially fewer environment interactions than
    our MoDem reproduction. Shaded regions denote 95\% normal-approximation
    confidence intervals.
    }
    \label{fig:metaworld_efficiency}
\end{figure}

Figure~\ref{fig:metaworld_efficiency} shows the corresponding learning curves. OneDemo improves rapidly from a strong initialization, exceeding $80\%$ average success after roughly $15$k interactions and approaching $90\%$ within $100$--$150$k interactions. Our MoDem reproduction improves more gradually over the same budget.

While we observe consistent improvement across all tasks with OneDemo, the
magnitude of the gain varies by task. This variation reflects the
inductive bias imposed by our trajectory-editing formulation. OneDemo makes
only a small number of high-level decisions at segment boundaries, while the action sequence within each segment does not change in response to any new states or observations. It is therefore most effective when task variation can be captured by a
small number of changes to segment timing and object-relative geometry. Gains are more modest on tasks such as Soccer and Stick Pull,
where success benefits from continual state-dependent correction.

OneDemo is also robust to source-demonstration choice; results using randomly selected successful demonstrations are provided in
Appendix~\ref{app:source_sensitivity}.

\paragraph{Ablations.}
The ablations in Table~\ref{tab:metaworld_results} show that temporal and geometric adaptation contribute differently across tasks. Learning segment timing alone increases average success from $67.3\%$ to $89.7\%$, accounting for most of the improvement of the full model ($92.7\%$). This indicates that,
when the demonstrated geometry remains reusable, adapting \emph{when} the trajectory transitions between semantic stages is often sufficient.

Residual geometric adaptation provides additional gains when translation-only anchor alignment is insufficient, most notably on Box Close ($66.8\% \rightarrow 83.6\%$) and Stick Pull ($45.2\% \rightarrow 56.8\%$). On more precision-sensitive tasks, geometric
adaptation can be essential: on DexMimicGen Threading, residual learning improves adaptation success substantially over segmentation alone; we provide this ablation in Appendix~\ref{app:threading_ablation}. While the relative importance of temporal and geometric adaptation varies across tasks, OneDemo does not need to assume which will dominate; both are optimized within the same structured editing space.

\subsection{OneDemo as a Demonstration Generator}
\label{sec:demo_aug_mod}
We evaluate OneDemo as a demonstration generator on DexMimicGen using
downstream imitation learning. The benchmark features long-horizon bimanual tasks across multiple robot embodiments, including both parallel-jaw grippers and dexterous hands.

For each task, we select one reliably replayable trajectory from the released dataset as our source demonstration. We do option-PPO training using only this demonstration and then
execute the learned policy under randomized task configurations. Successful
rollouts are retained until we collect 1,000 trajectories per task, matching the scale of the released DexMimicGen datasets.

We train image-based BC-RNN policies on the generated trajectories using the
DexMimicGen robomimic pipeline, with absolute end-effector target poses as
actions and all other training settings unchanged. Training is repeated over
three seeds; following the DexMimicGen evaluation protocol, we report the best
50-episode periodic evaluation for each seed and aggregate performance across
seeds.

\paragraph{Results.}
Table~\ref{tab:dexmimicgen_results} summarizes both the success rate of the
demonstration generator and the performance of policies trained on the
resulting data.
\begin{table*}[t]
    \centering
    \caption{{\textbf{Demonstration-generation and downstream-policy results
in DexMimicGen tasks (success rate, \%).}}
Trajectory edit success is the fraction of randomized task configurations
for which OneDemo successfully edits the source demonstration into a
successful trajectory. Bold indicates the best
downstream result per task.}
    \label{tab:dexmimicgen_results}

    \small
    \setlength{\tabcolsep}{4.5pt}
    \renewcommand{\arraystretch}{0.98}

    \resizebox{\linewidth}{!}{%
    \begin{tabular}{@{}llc@{\hspace{12pt}}ccc@{}}
    \toprule
    & &
    \multicolumn{1}{c}{\shortstack{\textbf{Trajectory Edit Success}}} &
    \multicolumn{3}{c}{\textbf{Downstream Policy Success}} \\
    \cmidrule(lr){3-3}\cmidrule(l){4-6}

    \textbf{Embodiment} &
    \textbf{Task} &
    \textbf{OneDemo} &
    \multicolumn{2}{c}{\textbf{DexMimicGen Data}~\citep{jiang2025dexmimicgen}} &
    \textbf{OneDemo Data} \\

    \cmidrule(lr){4-5}\cmidrule(l){6-6}
    & & &
    \textbf{BC-RNN} &
    \textbf{Diffusion Policy} &
    \textbf{BC-RNN} \\

        \midrule

        \multirow{3}{*}{\shortstack[l]{Bimanual Panda\\+ parallel-jaw grippers}}
          & Piece Assembly & 94.1  & $66.0 \pm 2.0$ & $80.7 \pm 0.9$ & $\mathbf{97.3 \pm 0.9}$ \\
          & Threading      & 100.0 & $68.0 \pm 1.6$ & $69.3 \pm 1.9$ & $\mathbf{100.0 \pm 0.0}$ \\
          & Transport      & 99.7  & $57.3 \pm 4.1$ & $83.3 \pm 0.9$ & $\mathbf{100.0 \pm 0.0}$ \\

        \cmidrule{1-6}

        \multirow{3}{*}{\shortstack[l]{Bimanual Panda\\+ dexterous hands}}
          & Box Cleanup    & 100.0 & $94.7 \pm 0.9$ & $92.0 \pm 4.3$ & $\mathbf{100.0 \pm 0.0}$ \\
          & Drawer Cleanup & 99.0  & $80.0 \pm 0.0$ & $76.0 \pm 0.0$ & $\mathbf{100.0 \pm 0.0}$ \\
          & Tray Lift      & 99.9  & $78.0 \pm 2.0$ & $88.7 \pm 0.9$ & $\mathbf{100.0 \pm 0.0}$ \\

        \cmidrule{1-6}

        \multirow{3}{*}{\shortstack[l]{GR-1 humanoid\\+ dexterous hands}}
          & Pouring        & 100.0 & $62.0 \pm 7.5$ & $79.3 \pm 0.9$ & $\mathbf{100.0 \pm 0.0}$ \\
          & Coffee         & 93.5  & $84.7 \pm 4.9$ & $77.3 \pm 0.9$ & $\mathbf{99.3 \pm 0.9}$ \\
          & Can Sorting    & 99.8  & $96.0 \pm 4.3$ & $97.3 \pm 0.9$ & $\mathbf{100.0 \pm 0.0}$ \\

        \midrule
        \multicolumn{2}{l}{\textbf{Mean}}
          & 98.4 & 76.3 & 82.7 & \textbf{99.6} \\

        \bottomrule
    \end{tabular}%
    }
\end{table*}
OneDemo achieves over 90\% trajectory edit success rate on all nine DexMimicGen tasks, with an average of 98.4\%. This reliability
extends to diverse manipulation regimes. For example,
\emph{Transport} requires a long-horizon asymmetric bimanual sequence involving
opening a container, manipulating multiple objects, and transferring an object
between arms, whereas \emph{Threading} requires precise coordinated motion
between the two manipulators. 

More importantly, our generated trajectories provide substantially better
supervision for downstream policy learning. At matched dataset scale, BC-RNN
trained on our demonstrations achieves near-perfect performance across all nine
tasks, averaging 99.6\% success compared with 76.3\% reported by DexMimicGen.
Our BC-RNN also outperforms the reported Diffusion Policy on every task,
despite using a simpler recurrent behavior-cloning architecture.

The learned BC-RNN policy also matches or exceeds the rollout success rate of
the data generator across all tasks, indicating that successful edited
trajectories can be consolidated into a more reliable downstream policy. We
hypothesize that the strong downstream performance arises from a combination of
\emph{broad task-configuration coverage} and \emph{behavioral consistency}.
Although OneDemo starts from only a single source demonstration, its generated
data spans a task-configuration distribution comparable to DexMimicGen while
exhibiting lower object-centric trajectory dispersion. Appendix~\ref{app:data_analysis}
provides a detailed analysis across all nine tasks.

\paragraph{Additional Results.}
We additionally evaluate OneDemo as a demonstration generator in a low-data regime on selected RoboTwin tasks. Diffusion Policy trained on only 50 OneDemo-generated trajectories per task improves average success from $28.5\%$ to $48.3\%$ across four bimanual tasks. Full results are provided in Appendix~\ref{app:robotwin}.

\section{Limitations and Future Work}

OneDemo is designed for tasks in which successful behavior can be adapted through
a sequence of object-centric trajectory edits. Extending it to richer perceptual inputs, such as images and point clouds, and to within-segment closed-loop correction could broaden its
applicability to dynamic and contact-sensitive settings. Another extension is to handle changes in interaction mode, such as switching
which arm manipulates an object when the original assignment becomes unreachable. The learned editor could also serve as a DAgger-style
data-collection policy, providing corrective demonstrations on states visited by
the downstream policy.

\section{Conclusion}
We introduced \textsc{OneDemo}, a reinforcement learning framework that turns a single successful
demonstration into a structured search space for adapting behavior to new task configurations.
Through learning when to transition between object-centric trajectory segments and how to geometrically
edit their motion, OneDemo avoids searching directly over long sequences of low-level
actions. The resulting trajectory editor serves both as a sample-efficient task policy and as a high-reliability
demonstration generator. Across multiple benchmarks, we show that this formulation enables strong policy learning with limited online interaction and produces high-quality training
data for downstream policies. Overall, our results suggest that a single successful demonstration can
serve not only as supervision, but as a reusable behavioral scaffold for searching both policies and
new training data.

\subsection*{AI Use Statement}
Generative AI tools were used to assist with writing refinement and figure presentation. An LLM was also used as part of the proposed method to automatically initialize the coarse semantic option sequence and associated task-relevant anchors from task metadata and downsampled telemetry, as described in~\Cref{par:option_init} and Appendix~\ref{app:option_initialization}. All AI-assisted outputs were reviewed by the authors, who take responsibility for the scientific content, experiments, analyses, results, and conclusions.




\bibliography{refereces}

@article{zhao2023learning,
  title={Learning fine-grained bimanual manipulation with low-cost hardware},
  author={Zhao, Tony Z and Kumar, Vikash and Levine, Sergey and Finn, Chelsea},
  journal={arXiv preprint arXiv:2304.13705},
  year={2023}
}

@article{chi2025diffusion,
  title={Diffusion policy: Visuomotor policy learning via action diffusion},
  author={Chi, Cheng and Xu, Zhenjia and Feng, Siyuan and Cousineau, Eric and Du, Yilun and Burchfiel, Benjamin and Tedrake, Russ and Song, Shuran},
  journal={The International Journal of Robotics Research},
  volume={44},
  number={10-11},
  pages={1684--1704},
  year={2025},
  publisher={Sage Publications Sage UK: London, England}
}

@article{ze20243d,
  title={3d diffusion policy: Generalizable visuomotor policy learning via simple 3d representations},
  author={Ze, Yanjie and Zhang, Gu and Zhang, Kangning and Hu, Chenyuan and Wang, Muhan and Xu, Huazhe},
  journal={arXiv preprint arXiv:2403.03954},
  year={2024}
}

@inproceedings{liu2025rdt,
  title={Rdt-1b: a diffusion foundation model for bimanual manipulation},
  author={Liu, Songming and Wu, Lingxuan and Li, Bangguo and Tan, Hengkai and Chen, Huayu and Wang, Zhengyi and Xu, Ke and Su, Hang and Zhu, Jun},
  booktitle={International Conference on Learning Representations},
  volume={2025},
  pages={29982--30009},
  year={2025}
}

@inproceedings{wang2024cyberdemo,
  title={Cyberdemo: Augmenting simulated human demonstration for real-world dexterous manipulation},
  author={Wang, Jun and Qin, Yuzhe and Kuang, Kaiming and Korkmaz, Yigit and Gurumoorthy, Akhilan and Su, Hao and Wang, Xiaolong},
  booktitle={2024 IEEE/CVF Conference on Computer Vision and Pattern Recognition (CVPR)},
  pages={17952--17963},
  year={2024},
  organization={IEEE}
}

@inproceedings{zhou2023nerf,
  title={Nerf in the palm of your hand: Corrective augmentation for robotics via novel-view synthesis},
  author={Zhou, Allan and Kim, Moo Jin and Wang, Lirui and Florence, Pete and Finn, Chelsea},
  booktitle={2023 IEEE/CVF Conference on Computer Vision and Pattern Recognition (CVPR)},
  pages={17907--17917},
  year={2023},
  organization={IEEE}
}

@article{mandlekar2023mimicgen,
  title={Mimicgen: A data generation system for scalable robot learning using human demonstrations},
  author={Mandlekar, Ajay and Nasiriany, Soroush and Wen, Bowen and Akinola, Iretiayo and Narang, Yashraj and Fan, Linxi and Zhu, Yuke and Fox, Dieter},
  journal={arXiv preprint arXiv:2310.17596},
  year={2023}
}

@inproceedings{yuan2026demograsp,
  title={Demograsp: Universal dexterous grasping from a single demonstration},
  author={Yuan, Haoqi and Huang, Ziye and Wang, Ye and Mao, Chuan and Xu, Chaoyi and Lu, Zongqing},
  booktitle={International Conference on Learning Representations},
  volume={2026},
  pages={64609--64633},
  year={2026}
}

@inproceedings{zhu2025dexflywheel,
 author = {Zhu, Kefei and Bai, Fengshuo and Xiang, YuanHao and Cai, Yishuai and Chen, Xinglin and Li, Ruochong and Wang, Xingtao and Dong, Hao and Yang, Yaodong and Fan, Xiaopeng and Chen, Yuanpei},
 booktitle = {Advances in Neural Information Processing Systems},
 doi = {10.52202/085713-0150},
 editor = {D. Belgrave and C. Zhang and H. Lin and R. Pascanu and P. Koniusz and M. Ghassemi and N. Chen},
 pages = {4236--4265},
 publisher = {Curran Associates, Inc.},
 title = {DexFlyWheel: A Scalable and Self-improving Data Generation Framework for Dexterous Manipulation},
 url = {https://proceedings.neurips.cc/paper_files/paper/2025/file/066c5542795287822f4f076cf330e5a2-Paper-Conference.pdf},
 volume = {38, Main Conference},
 year = {2025}
}

@article{osa2018algorithmic,
  title={An algorithmic perspective on imitation learning},
  author={Osa, Takayuki and Pajarinen, Joni and Neumann, Gerhard and Bagnell, J Andrew and Abbeel, Pieter and Peters, Jan},
  journal={Foundations and Trends{\textregistered} in Robotics},
  volume={7},
  number={1-2},
  pages={1--179},
  year={2018},
  publisher={Emerald Publishing Limited}
}

@article{mandlekar2021matters,
  title={What matters in learning from offline human demonstrations for robot manipulation},
  author={Mandlekar, Ajay and Xu, Danfei and Wong, Josiah and Nasiriany, Soroush and Wang, Chen and Kulkarni, Rohun and Fei-Fei, Li and Savarese, Silvio and Zhu, Yuke and Mart{\'\i}n-Mart{\'\i}n, Roberto},
  journal={arXiv preprint arXiv:2108.03298},
  year={2021}
}

@inproceedings{jiang2025dexmimicgen,
  title={Dexmimicgen: Automated data generation for bimanual dexterous manipulation via imitation learning},
  author={Jiang, Zhenyu and Xie, Yuqi and Lin, Kevin and Xu, Zhenjia and Wan, Weikang and Mandlekar, Ajay and Fan, Linxi Jim and Zhu, Yuke},
  booktitle={2025 IEEE International Conference on Robotics and Automation (ICRA)},
  pages={16923--16930},
  year={2025},
  organization={IEEE}
}

@inproceedings{ross2011reduction,
  title={A reduction of imitation learning and structured prediction to no-regret online learning},
  author={Ross, St{\'e}phane and Gordon, Geoffrey and Bagnell, Drew},
  booktitle={Proceedings of the fourteenth international conference on artificial intelligence and statistics},
  pages={627--635},
  year={2011},
  organization={JMLR Workshop and Conference Proceedings}
}

@article{mandlekar2020human,
  title={Human-in-the-loop imitation learning using remote teleoperation},
  author={Mandlekar, Ajay and Xu, Danfei and Mart{\'\i}n-Mart{\'\i}n, Roberto and Zhu, Yuke and Fei-Fei, Li and Savarese, Silvio},
  journal={arXiv preprint arXiv:2012.06733},
  year={2020}
}

@inproceedings{nair2018overcoming,
  title={Overcoming exploration in reinforcement learning with demonstrations},
  author={Nair, Ashvin and McGrew, Bob and Andrychowicz, Marcin and Zaremba, Wojciech and Abbeel, Pieter},
  booktitle={2018 IEEE international conference on robotics and automation (ICRA)},
  pages={6292--6299},
  year={2018},
  organization={IEEE}
}

@article{rajeswaran2017learning,
  title={Learning complex dexterous manipulation with deep reinforcement learning and demonstrations},
  author={Rajeswaran, Aravind and Kumar, Vikash and Gupta, Abhishek and Vezzani, Giulia and Schulman, John and Todorov, Emanuel and Levine, Sergey},
  journal={arXiv preprint arXiv:1709.10087},
  year={2017}
}

@article{vecerik2017leveraging,
  title={Leveraging demonstrations for deep reinforcement learning on robotics problems with sparse rewards},
  author={Vecerik, Mel and Hester, Todd and Scholz, Jonathan and Wang, Fumin and Pietquin, Olivier and Piot, Bilal and Heess, Nicolas and Roth{\"o}rl, Thomas and Lampe, Thomas and Riedmiller, Martin},
  journal={arXiv preprint arXiv:1707.08817},
  year={2017}
}

@inproceedings{johannink2019residual,
  title={Residual reinforcement learning for robot control},
  author={Johannink, Tobias and Bahl, Shikhar and Nair, Ashvin and Luo, Jianlan and Kumar, Avinash and Loskyll, Matthias and Ojea, Juan Aparicio and Solowjow, Eugen and Levine, Sergey},
  booktitle={2019 international conference on robotics and automation (ICRA)},
  pages={6023--6029},
  year={2019},
  organization={IEEE}
}

@InProceedings{Garrett2025skillmimic,
  title = 	 {SkillMimicGen: Automated Demonstration Generation for Efficient Skill Learning and Deployment},
  author =       {Garrett, Caelan Reed and Mandlekar, Ajay and Wen, Bowen and Fox, Dieter},
  booktitle = 	 {Proceedings of The 8th Conference on Robot Learning},
  pages = 	 {2750--2790},
  year = 	 {2025},
  editor = 	 {Agrawal, Pulkit and Kroemer, Oliver and Burgard, Wolfram},
  volume = 	 {270},
  series = 	 {Proceedings of Machine Learning Research},
  month = 	 {06--09 Nov},
  publisher =    {PMLR},
  url = 	 {https://proceedings.mlr.press/v270/garrett25a.html}
}

@inproceedings{bacon2017option,
  title={The option-critic architecture},
  author={Bacon, Pierre-Luc and Harb, Jean and Precup, Doina},
  booktitle={Proceedings of the AAAI conference on artificial intelligence},
  volume={31},
  number={1},
  year={2017}
}

@article{sutton1999between,
  title={Between MDPs and semi-MDPs: A framework for temporal abstraction in reinforcement learning},
  author={Sutton, Richard S and Precup, Doina and Singh, Satinder},
  journal={Artificial intelligence},
  volume={112},
  number={1-2},
  pages={181--211},
  year={1999},
  publisher={Elsevier}
}

@inproceedings{yu2020meta,
  title={Meta-world: A benchmark and evaluation for multi-task and meta reinforcement learning},
  author={Yu, Tianhe and Quillen, Deirdre and He, Zhanpeng and Julian, Ryan and Hausman, Karol and Finn, Chelsea and Levine, Sergey},
  booktitle={Conference on robot learning},
  pages={1094--1100},
  year={2020},
  organization={PMLR}
}

@article{hansen2022modem,
  title={Modem: Accelerating visual model-based reinforcement learning with demonstrations},
  author={Hansen, Nicklas and Lin, Yixin and Su, Hao and Wang, Xiaolong and Kumar, Vikash and Rajeswaran, Aravind},
  journal={arXiv preprint arXiv:2212.05698},
  year={2022}
}

@article{chen2025robotwin,
  title={Robotwin 2.0: A scalable data generator and benchmark with strong domain randomization for robust bimanual robotic manipulation},
  author={Chen, Tianxing and Chen, Zanxin and Chen, Baijun and Cai, Zijian and Liu, Yibin and Li, Zixuan and Liang, Qiwei and Lin, Xianliang and Ge, Yiheng and Gu, Zhenyu and others},
  journal={arXiv preprint arXiv:2506.18088},
  year={2025}
}

@article{schulman2017proximal,
  title={Proximal policy optimization algorithms},
  author={Schulman, John and Wolski, Filip and Dhariwal, Prafulla and Radford, Alec and Klimov, Oleg},
  journal={arXiv preprint arXiv:1707.06347},
  year={2017}
}
\bibliographystyle{iclr2027_conference}

\appendix
\section{Appendix}

\subsection{Background on the Options Framework}
\label{app:options}

The options framework~\citep{sutton1999between} extends a Markov
decision process (MDP) with temporally extended actions. Consider an MDP
with state space $\mathcal{S}$ and primitive action space $\mathcal{A}$.
A Markovian option $\omega \in \Omega$ is defined as a tuple
\[
    \omega = (I_\omega, \pi_\omega, \beta_\omega),
\]
where $I_\omega \subseteq \mathcal{S}$ is the \emph{initiation set},
$\pi_\omega(a \mid s)$ is the \emph{intra-option policy}, and
$\beta_\omega(s) \in [0,1]$ is the \emph{termination function}.
The initiation set specifies the states in which the option may be started, the intra-option policy determines the primitive actions executed while the option is active, and the termination function gives the probability that the option terminates upon reaching a state.

Given a set of options $\Omega$, a separate \emph{policy over options}
\[
    \pi_\Omega(\omega \mid s)
\]
determines which option to initiate. Under the standard
\emph{call-and-return} execution model, an option is sampled according
to $\pi_\Omega$ and its intra-option policy is then followed until the
option terminates. More precisely, if $\omega_t$ is active in state
$s_t$, a primitive action is sampled according to the intra-option policy as
\[
    a_t \sim \pi_{\omega_t}(\cdot \mid s_t).
\]
After the environment transitions to $s_{t+1}$, the active option
terminates with probability $\beta_{\omega_t}(s_{t+1})$. If it does not
terminate, execution continues under the same option; otherwise, a new
option is selected according to
$\pi_\Omega(\cdot \mid s_{t+1})$.

Thus, the options framework introduces two levels of decision making:
the policy over options determines \emph{which intra-option policy to execute}, while the intra-option policy determines
\emph{the actual primitive action carried out at each time step}.
The termination function determines when to sample a new option. Because an option may persist for a variable number
of primitive time steps, an MDP augmented with options naturally induces
a semi-Markov decision process (SMDP) ~\citep{sutton1999between} at the option-decision
level.

The Option-Critic architecture~\citep{bacon2017option} shows that the
intra-option policies and termination functions can be optimized directly from task return. These components are learned jointly with a policy over options, whose learning rule can be chosen separately.

\subsection{Choice of Anchor Alignment}
\label{app:anchor_alignment}

For each semantic option, OneDemo uses translation-only alignment of the
associated anchor as its deterministic geometric initialization:
\[
B_i
=
\operatorname{Trans}(p_i-p_i^{\mathrm{src}}).
\]
We deliberately do not directly align the full source and live anchor poses.

There are two motivations for this choice. First, changes in anchor orientation
do not always imply that the demonstrated end-effector motion should undergo the
same rotation. This is particularly common for objects with rotational
symmetries, where multiple object orientations correspond to equivalent
manipulation geometry. Directly inheriting the full anchor rotation can
therefore introduce unnecessary changes to an otherwise reusable trajectory.

Second, full pose alignment can rotate the demonstrated trajectory into
kinematically difficult or infeasible end-effector configurations. Existing
trajectory-generation systems such as MimicGen~\citep{mandlekar2023mimicgen} and DexMimicGen~\citep{jiang2025dexmimicgen} mitigate this
issue in part by drawing from multiple source demonstrations and selecting
source segments that are compatible with the current task configuration. OneDemo instead assumes
only a single source demonstration.

We therefore use translation-only alignment as a conservative initialization
and let the learned residual
\[
\Delta T_i \in SE(3)
\]
supply any additional translational or rotational correction required by the
task. This preserves the demonstrated motion whenever it remains reusable while
still allowing the policy to depart from the initialization when task success
requires stronger geometric adaptation.

\subsection{Automatic Semantic Option Initialization}
\label{app:option_initialization}

OneDemo requires an ordered sequence of task-relevant anchors that defines the
semantic option sequence $\Omega=(\omega_1,\ldots,\omega_K)$. Each option
$\omega_i$ is associated with an anchor object or task frame, whose source and
live poses determine the baseline geometric transformation used during
trajectory editing.

In principle, this structure could be specified manually. In our
implementation, we initialize it automatically using an LLM. The model receives
task metadata and downsampled telemetry from the successful source
demonstration and proposes an ordered semantic decomposition together with the
anchor associated with each option. This task requires only coarse semantic
interpretation of the demonstration; the LLM does not infer precise temporal
boundaries, contact dynamics, or low-level robot actions.

The predicted option sequence need not be minimal. For example, the LLM may
separate reaching and grasping into two options even when both use the same
object anchor. Such redundancy is tolerated by the learned duration policy,
which can assign near-minimal duration to unnecessary options. The LLM
therefore provides only a semantic scaffold, while OneDemo learns the temporal
segmentation and geometric adaptation from task return.

\paragraph{Prompt.}
The exact prompt used in our DexMimicGen experiments is reproduced below.
The bracketed metadata and telemetry fields are populated automatically for
each source demonstration.

\begin{lstlisting}[
  basicstyle=\ttfamily\scriptsize,
  breaklines=true,
  breakatwhitespace=true,
  columns=fullflexible,
  keepspaces=true,
  showstringspaces=false,
  breakindent=1em
]
You are an expert in robosuite and DexMimicGen demonstration composition.

Analyze the successful bimanual demonstration below and generate a COMPLETE,
DIRECTLY LOADABLE DexMimicGen experiment JSON. Infer the semantic option
program from telemetry; no existing generation JSON is provided.

### Hand and telemetry conventions
- `right_*` telemetry is `right_hand`; `left_*` telemetry is `left_hand`.
- Infer each hand's ordered semantic options, their anchor objects, and their
  coordination. Do not infer frame indices, durations, or temporal boundaries;
  Option PPO learns option transitions.
- Use object poses, EEF motion, gripper motion, and termination signals as
  semantic evidence. Associate a termination signal with the option whose
  completion it describes, but do not estimate when the option ends.
- Treat signal ownership as a causal question. At each `signals_rising` event,
  use `eef_object_distances`, object motion, and gripper motion to identify
  which hand is actually manipulating the relevant object. Assign a global
  task signal to that hand's option; assign it to both hands only when both
  are still causally engaged in the same coordinated operation at activation.
- Respect explicit hand semantics in signal names such as `_right` and `_left`
  unless the telemetry clearly contradicts them.
- `object_ref` MUST exactly match one name in `object_refs_in_demo`. Never use
  a robot, EEF, TCP, gripper, action channel, or invented alias.
- A non-null `subtask_term_signal` MUST exactly match one name in
  `subtask_term_signals`. Do not invent signals.
- Associate each available termination signal with the option whose completed
  semantics it describes. This is a label association, not boundary inference.
  The final option for each hand MUST have `subtask_term_signal: null`.
- Merge approach and grasp preparation into an acquire phase anchored to the
  manipulated object. After acquisition, change the anchor to the receiving
  object, fixture, or goal that governs placement/insertion.
- Keep grasp, lift, carry, receive, and handoff options anchored to the
  manipulated or transferred object. Use a destination, receptacle, fixture,
  or goal as the anchor only for an option whose semantics are governed by
  placement, insertion, alignment, or release at that destination.
- Apply a conditional destination selector only to destination-anchored
  options. Do not replace the manipulated-object anchor of grasp, lift, carry,
  receive, or handoff options merely because the eventual destination is
  conditional.
- Preserve temporal order. Use concise snake_case phase names such as
  `acquire_object`, `place_object`, or `insert_object`.
- When both hands participate in the same coupled phase, give both
  corresponding phase entries the exact same snake_case `coordination_group`.
  Omit `coordination_group` for independent phases.
- If an option's live anchor depends on an episode condition, include
  `live_object_ref` with an exact condition name and a complete mapping from
  every observed condition value to an object in `object_refs_in_demo`.
  Infer this mapping from the condition-stratified semantic examples.
  Keep `object_ref` equal to the anchor used by the selected source episode.
  Omit `live_object_ref` for unconditional options.
- Before returning JSON, silently audit every option: verify that its anchor
  matches its semantic verb, every signal is assigned to the hand or
  coordinated hands causally responsible at activation, and conditional
  selectors appear only where the live anchor truly changes with the condition.
- Do not add tuning fields such as action_noise, offsets, interpolation steps,
  or selection strategies.

### Exact output contract
Return JSON only. Do not add `trajectory_status`, `task_name`, `evidence`,
comments, markdown, or any other fields. Copy the fixed boilerplate values
below exactly. Replace only angle-bracket placeholders and dynamically
populate the ordered phase mappings.

{
  "name": "<suggested_config_name from HDF5 metadata>",
  "type": "robosuite",
  "experiment": {
    "name": "demo",
    "source": {
      "dataset_path": null,
      "filter_key": null,
      "n": null,
      "start": null
    },
    "generation": {
      "path": null,
      "guarantee": false,
      "keep_failed": true,
      "num_trials": 10,
      "select_src_per_subtask": false,
      "transform_first_robot_pose": false,
      "interpolate_from_last_target_pose": true
    },
    "task": {
      "name": "<env_name from HDF5 metadata>",
      "robot": null,
      "gripper": null,
      "env_meta_update_kwargs": {},
      "interface": "<env_interface_name from HDF5 metadata>",
      "interface_type": "<env_interface_type from HDF5 metadata>"
    },
    "max_num_failures": 50,
    "render_video": true,
    "num_demo_to_render": 50,
    "num_fail_demo_to_render": 50,
    "log_every_n_attempts": 50,
    "seed": 1
  },
  "obs": {
    "collect_obs": true,
    "camera_names": [],
    "camera_height": 84,
    "camera_width": 84
  },
  "task": {
    "task_spec": {
      "right_hand": {
        "<ordered_phase_name>": {
          "object_ref": "<exact object_refs_in_demo name>",
          "subtask_term_signal":
              "<exact subtask_term_signals name or null>",
          "coordination_group":
              "<shared group only when coordinated; otherwise omit this field>",
          "live_object_ref": {
            "condition": "<exact observed condition name>",
            "cases": {
              "<observed value>": "<exact object_refs_in_demo name>"
            }
          }
        }
      },
      "left_hand": {
        "<ordered_phase_name>": {
          "object_ref": "<exact object_refs_in_demo name>",
          "subtask_term_signal":
              "<exact subtask_term_signals name or null>",
          "coordination_group":
              "<shared group only when coordinated; otherwise omit this field>",
          "live_object_ref": {
            "condition": "<exact observed condition name>",
            "cases": {
              "<observed value>": "<exact object_refs_in_demo name>"
            }
          }
        }
      }
    }
  }
}

[TRAIN CONFIG METADATA START]
<automatically populated training configuration metadata>
[TRAIN CONFIG METADATA END]

[HDF5 METADATA START]
<automatically populated HDF5 metadata>
[HDF5 METADATA END]

[CONDITION-STRATIFIED SEMANTIC EXAMPLES START]
<automatically populated condition-stratified semantic examples>
[CONDITION-STRATIFIED SEMANTIC EXAMPLES END]

[DOWNSAMPLED TELEMETRY START]
<automatically populated trajectory telemetry>
[DOWNSAMPLED TELEMETRY END]
\end{lstlisting}

\subsection{Source Demo Selection}
\label{app:source_sensitivity}
The least-motion demonstrations used in our main experiments are not particularly favorable source trajectories: several randomly selected sources have substantially higher naive success. Nevertheless, OneDemo substantially reduces sensitivity to source choice. The performance range across the four sources decreases from \(34.0\) to \(14.0\) percentage points on Box Close and from \(94.4\) to \(14.4\) points on Basketball after training. In the most extreme case, a Basketball source with only \(3.2\%\) naive success reaches \(98.0\%\) best success after adaptation.

\begin{table}[H]
\centering
\caption{
Sensitivity to source-demonstration selection on Meta-World Box Close and
Basketball. We compare the least-motion demonstration used in the main
experiments with three randomly selected successful source demonstrations.
Values are mean $\pm$ standard deviation over five training seeds.
Naive denotes performance before RL adaptation, and OneDemo denotes best
checkpoint performance, consistent with ~\Cref{tab:metaworld_results}.
}
\label{tab:source_demo_sensitivity}
\begin{tabular}{llcc}
\toprule
Task & Source Demonstration & Naive & OneDemo \\
\midrule
Box Close
& Least-motion demo & $56.8 \pm 8.1$  & $83.6 \pm 14.5$ \\
& Random demo 1      & $88.4 \pm 1.5$  & $97.2 \pm 1.6$ \\
& Random demo 2      & $72.8 \pm 5.2$  & $97.6 \pm 1.5$ \\
& Random demo 3       & $90.8 \pm 5.2$  & $96.4 \pm 1.5$ \\
\midrule
Basketball
& Least-motion demo & $67.6 \pm 15.7$ & $84.4 \pm 5.6$ \\
& Random demo 1        & $3.2 \pm 2.4$   & $98.0 \pm 2.5$ \\
& Random demo 2       & $97.6 \pm 2.3$  & $98.8 \pm 1.6$ \\
& Random demo 3       & $78.0 \pm 3.3$  & $88.0 \pm 2.8$ \\
\bottomrule
\end{tabular}
\end{table}

\subsection{Implementation Details}
\label{app:implementation}

\paragraph{Option-PPO Architecture.}
Across all benchmarks, we use the same decision-level actor--critic
architecture for Option-PPO. At each option boundary, the policy observation
is encoded by a two-layer MLP with 256 hidden units per layer and Tanh
activations. A categorical duration head predicts a distribution over discrete
option-duration bins; in bimanual settings, each arm has a separate duration
head. Duration bin $b$ advances the source trajectory by $(b+1)c$ steps, where
$c$ is the benchmark-specific commitment interval.

A linear value head predicts the scalar state value. The residual policy
concatenates the shared trunk features with the raw policy observation, passes
the result through an additional 256-unit Tanh layer, and predicts the mean of
a diagonal Gaussian residual distribution. The residual policy uses a learned,
state-independent log standard deviation for each action dimension.

We optimize the duration and residual policies using clipped PPO. Training
alternates between segmentation updates, which optimize the duration policy,
and warp updates, which optimize the residual policy; the inactive component is
frozen during each stage. During segmentation updates, the residual action is
set to the mean of the current Gaussian residual policy.

\paragraph{PPO Optimization.}
Table~\ref{tab:shared_ppo_hparams} summarizes the shared setup parameters across
benchmarks. The network architecture is shared across benchmarks, while the residual action representation, commitment interval, optimization
schedule, and rollout budget reflect the corresponding control interface.
Table~\ref{tab:benchmark_ppo_hparams} summarizes the principal differences.
\begin{table}[t]
    \centering
    \caption{Hyperparameters shared by Option-PPO across benchmarks.}
    \label{tab:shared_ppo_hparams}
    \small
    \setlength{\tabcolsep}{5pt}
    \begin{tabular}{lc}
        \toprule
        \textbf{Hyperparameter} & \textbf{Value} \\
        \midrule
        Policy/value trunk & 2-layer MLP \\
        Hidden width & 256 \\
        Activation & Tanh \\
        Optimizer & Adam \\
        Adam $\epsilon$ & $10^{-5}$ \\
        PPO clipping coefficient & $0.2$ \\
        Value-loss coefficient & $0.5$ \\
        Maximum gradient norm & $0.5$ \\
        Advantage normalization & per batch \\
        Discount factor / GAE & not used \\
        Return & terminal success \\
        Residual distribution & diagonal Gaussian \\
        Residual log std. & learned, state-independent \\
        \bottomrule
    \end{tabular}
\end{table}

\begin{table*}[t]
    \centering
    \caption{Benchmark-specific Option-PPO settings.}
    \label{tab:benchmark_ppo_hparams}
    \small
    \setlength{\tabcolsep}{5pt}
    \resizebox{\linewidth}{!}{%
    \begin{tabular}{lccc}
        \toprule
        \textbf{Hyperparameter}
        & \textbf{Meta-World}
        & \textbf{DexMimicGen}
        & \textbf{RoboTwin} \\
        \midrule
        Learning rate
            & $3\times10^{-4}$ & $3\times10^{-4}$ & $3\times10^{-4}$ \\
        LR schedule
            & constant & constant & linear decay \\
        PPO epochs/update
            & 4 & 4 & 4 \\
        Minibatches/update
            & 2 & 32 & 2 \\
        Episodes per PPO update
            & 100--188, task-dependent
            & 2,048
            & 60 or 120, task-dependent \\
        Number of PPO updates
            & 16
            & 100
            & 10 or 20, task-dependent \\
        Commitment interval $c$
            & 10 & 5 & 20 \\
        Residual action
            & 3-D translation
            & 6-DoF translation + rotation
            & 6-DoF translation + rotation \\
        Residual frame
            & target XYZ offset
            & anchor frame
            & anchor frame \\
        \bottomrule
    \end{tabular}%
    }
\end{table*}

\paragraph{MoDem Baseline.}
\label{par:modem_baseline}
We use MoDem~\citep{hansen2022modem} as a reference baseline for the
Meta-World experiments. MoDem is a demonstration-guided reinforcement
learning method designed for sample-efficient learning under sparse rewards. It uses demonstrations to initialize the policy and guide subsequent online reinforcement learning, but does not constrain exploration to structured edits of a demonstrated trajectory. We therefore use our local reproduction of MoDem
to contextualize the interaction scale required when demonstrations guide sparse-reward policy learning without imposing the reduced trajectory-editing action space from OneDemo.

\subsection{Geometric Adaptation on DexMimicGen Threading}
\label{app:threading_ablation}
To further isolate the role of geometric residual adaptation, we compare the segmentation-only and full OneDemo policies on DexMimicGen Threading. Threading requires precise coordinated bimanual motion, making translation-only anchor alignment substantially less reliable than on many Meta-World tasks.

Under the same $20$M environment-interaction budget, the segmentation-only variant reaches $32.1\%$ success, whereas the full OneDemo policy with learned geometric residuals reaches $81.8\%$. This large gap shows that temporal adaptation alone is insufficient on this task: successful execution requires substantial geometric correction of the demonstrated trajectory.

\subsection{Generated Data Analysis}
\label{app:data_analysis}

To better understand the strong downstream imitation-learning performance of
OneDemo-generated data, we compare its task-configuration coverage and
behavioral variation against the released DexMimicGen demonstrations.

Despite using only a single source demonstration, OneDemo preserves essentially
the full task-configuration diversity of the released DexMimicGen datasets.
Across all nine tasks, the OneDemo-to-DexMimicGen task-configuration coverage
ratio ranges from $0.990\times$ to $1.016\times$, with a mean of $1.003\times$.
Here, coverage is measured from the 95\% span of each randomized object-pose
dimension and aggregated across dimensions within each task.

We next analyze variation in the demonstrated robot behavior after factoring
out this task-configuration variation. We express each end-effector trajectory
in its corresponding object-anchor frame and temporally normalize each semantic
phase. For each phase, we measure the RMS translational and rotational deviation
from the cross-demonstration mean trajectory, and average this dispersion over
normalized time and phases.

\begin{figure}[t]
    \centering
    \includegraphics[width=\linewidth]{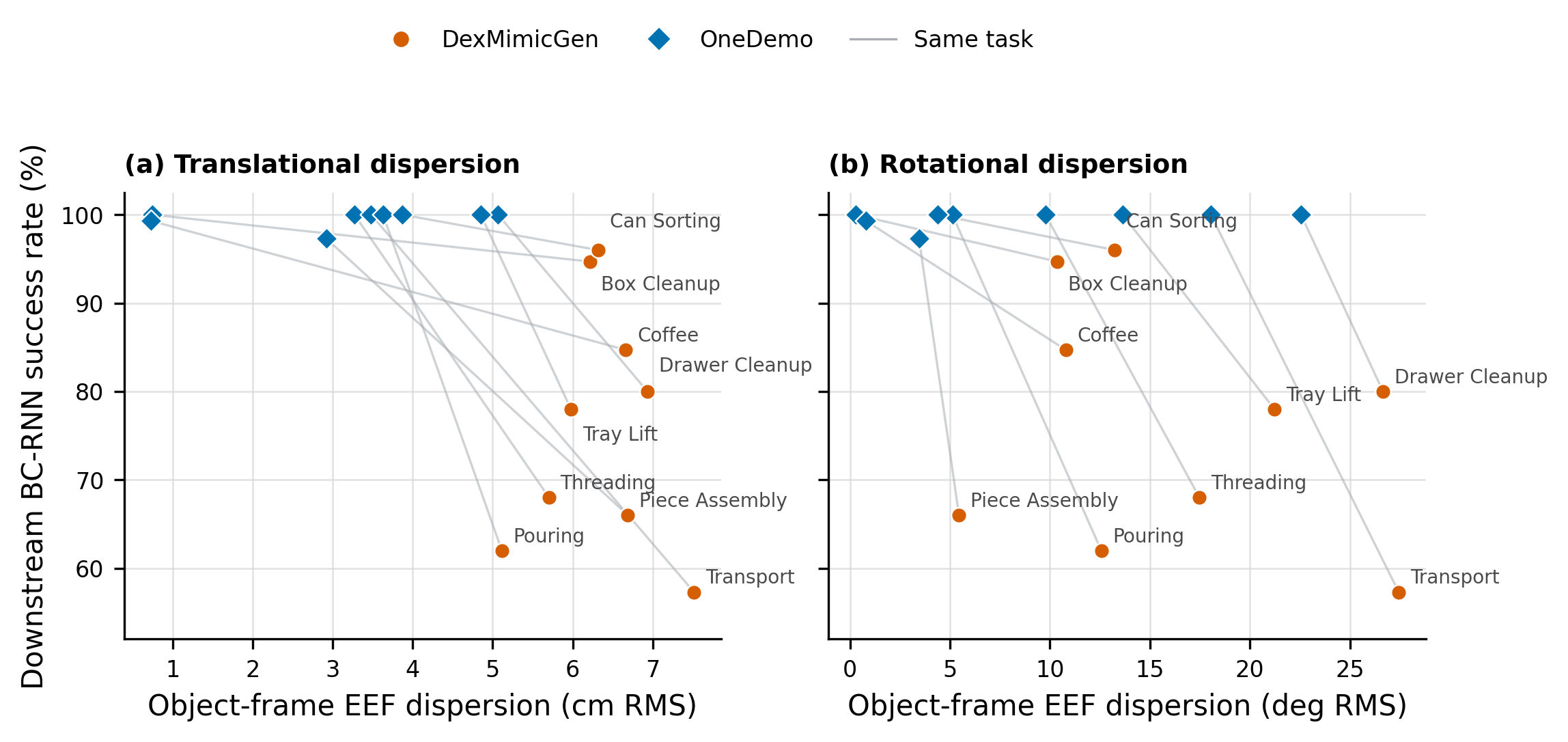}
    \caption{
    Object-centric behavioral dispersion and downstream imitation-learning
    performance across all nine DexMimicGen tasks.
    Each gray line connects DexMimicGen and OneDemo data for the same task.
    The left and right panels show mean phase-wise translational and rotational
    end-effector trajectory dispersion, respectively.
    OneDemo consistently produces lower behavioral dispersion while improving
    downstream BC-RNN success.
    }
    \label{fig:dispersion_bc}
\end{figure}

Figure~\ref{fig:dispersion_bc} relates this object-centric behavioral dispersion
to downstream BC-RNN performance. Each line connects DexMimicGen and OneDemo
for the same task. Across all nine tasks, OneDemo-generated data exhibits lower
average translational and rotational dispersion while the corresponding
downstream BC-RNN also achieves higher success.

These results are consistent with the hypothesis that OneDemo
achieves a favorable combination of \emph{coverage} and \emph{consistency}:
starting from a single source demonstration, it generates trajectories across
a broad range of task configurations while largely preserving a consistent
object-relative manipulation strategy. This provides a plausible explanation
for why the generated demonstrations can be more easily consolidated into an imitation-learning policy.

\subsection{Robotwin}
\label{app:robotwin}
RoboTwin provides a challenging setting for OneDemo because many tasks admit
multiple behavior modes that cannot be obtained through geometric adaptation of
a single trajectory. In particular, a task may require different arm assignments
depending on the scene configuration, whereas OneDemo currently adapts behavior
within a fixed interaction mode rather than switching between modes.

For \emph{Handover Block} and \emph{Pick Dual Bottles}, a single behavior mode
was sufficient, so we use one source demonstration to train Option-PPO and
generate trajectories. \emph{Beat Block Hammer} and \emph{Place Empty Cup}
require two behavior modes, corresponding to performing the manipulation with
the left or right arm. For these tasks, we select one source demonstration per
mode and treat the two modes as separate adaptation problems, training an
independent trajectory editor for each. The generated trajectories are then
used to train the downstream Diffusion Policy. Results are shown in
Table~\ref{tab:robotwin_results}.

\begin{table}[H]
\centering
\caption{
RoboTwin results on four bimanual tasks.
\textit{Trajectory edit success} is the fraction of randomized task configurations
for which OneDemo successfully edits the source demonstration into a
successful trajectory.
\textit{DP Base} denotes our reproduced Diffusion Policy performance using
RoboTwin's released demonstrations, while \textit{DP OneDemo} denotes Diffusion
Policy performance using OneDemo-generated demonstrations.
Numbers in parentheses indicate absolute percentage-point improvement over
\textit{DP Base}.
}
\label{tab:robotwin_results}
\setlength{\tabcolsep}{6pt}
\begin{tabular}{lccc}
\toprule
\textbf{Task}
& \textbf{Trajectory Edit Success}
& \textbf{DP Base}
& \textbf{DP OneDemo} \\
\midrule
Beat Block Hammer & 96.7 & 42 & \textbf{56}~(+14) \\
Handover Block    & 98.3 & 15 & \textbf{20}~(+5)  \\
Pick Dual Bottles & 98.3 & 25 & \textbf{52}~(+27) \\
Place Empty Cup   & 96.7 & 32 & \textbf{65}~(+33) \\
\midrule
\textbf{Average}  & 97.5 & 28.5 & \textbf{48.3}~(+19.8) \\
\bottomrule
\end{tabular}
\end{table}
\end{document}